\documentclass[runningheads]{llncs}
\usepackage[T1]{fontenc}
\usepackage{booktabs}
\usepackage{pifont}

\usepackage{hyperref}

\usepackage{amssymb}

\newcommand{\cmark}{\ding{51}}
\newcommand{\xmark}{\ding{55}}
\usepackage[table]{xcolor}
\definecolor{motion}{RGB}{235,245,255}   
\definecolor{vision}{RGB}{245,245,245}   
\definecolor{fusion}{RGB}{235,255,235}   

\usepackage{cite}

\usepackage{graphicx,verbatim}
\begin{document}
\title{Motion-Conditioned Multi-View Fusion for Myocardial Infarction Localization from Echocardiography}

\titlerunning{Motion-Conditioned Multi-View Fusion}
%

\author{
Guang Yang\inst{1}\thanks{Corresponding author.}
\and
Wentian Xu\inst{1}
\and
Siyu Wang\inst{1}
\and
Betty Raman\inst{2}
\and 
Lei Li\inst{3}
\and
Vicente Grau\inst{1}
}

\authorrunning{G. Yang et al.}

\institute{
Department of Engineering Science,
University of Oxford,
Oxford, United Kingdom\\
\email{guang.yang@eng.ox.ac.uk}
\and
Radcliffe Department of Medicine,
University of Oxford,
Oxford, United Kingdom
\and
Department of Biomedical Engineering,
National University of Singapore,
Singapore
}

\maketitle              
\begin{abstract}
Myocardial infarction (MI) remains a leading cause of mortality worldwide. Echocardiography (Echo) is a widely available modality for MI assessment, where regional wall motion abnormality is a key indicator. Prior learning based methods for myocardial motion analysis often use handcrafted descriptors or densely supervised estimation, but the need for extensive annotation limits applicability.  Foundation models have recently improved vision-based Echo analysis; however, most methods operate on single views and segment-level localization remains unreliable under view-dependent ambiguity, especially in apical views. To address this, we propose MCF-Net, a novel motion-guided multi-view fusion framework that fuses myocardial motion cues with foundation model representations to localize infarction. Visual features are extracted using EchoPrime, a pretrained Echo foundation model shared across dual views. Cardiac motion is modeled with extremely sparse supervision: a single annotated template frame is transferred across videos to initialize point tracking, avoiding dense labels. Motion-derived segment-aware soft masks provide coarse spatial priors that selectively enhance features for challenging myocardial segments. A motion-conditioned fusion mechanism then integrates motion and vision across views, refining predictions without overriding strong appearance cues. On segment-level MI localization, MCF-Net achieves 72.4\% F1 and 84.9\% accuracy, outperforming state-of-the-art motion-only, vision-only, and fusion baselines.


\end{abstract}
\section{Introduction}
Myocardial infarction (MI) refers to irreversible myocardial damage caused by coronary artery occlusion. The associated mechanical dysfunction often results in abnormal myocardial motion \cite{zeidan2024myocardial}. Related contrast-free work in cine MRI has
also used cardiac motion to reconstruct patient-specific infarct geometry
\cite{lyu2025personalized}, supporting the  value of cardiac mechanics for infarct assessment. Echocardiography (Echo), with its high temporal resolution, is routinely used to assess MI via regional wall motion abnormality (RWMA). In clinical practice, MI assessment integrates multiple acquisition views; the assessment is most commonly made in apical four- and two-chamber (A4C/A2C) views, which provide different coverage of the left ventricle. Following the AHA 17-segment model \cite{cerqueira2002standardized}, each apical view maps to six distinct myocardial segments. However, Echo often suffers from limited image quality and view-dependent ambiguity, making RWMA assessment highly dependent on clinician expertise \cite{porter2018clinical} and hindering scalable, reproducible MI localization.

Most automated approaches model RWMA through explicit motion analysis. Optical flow based methods estimate pixel wise motion without anatomical labels but are sensitive to noise and imaging artifacts, often yielding physically implausible motion fields \cite{suhling2005myocardial,ortiz2023optical}. Image registration methods estimate dense deformation between frames, yet their performance degrades without accurate anatomical supervision \cite{balakrishnan2019voxelmorph,yang2025contrast,bi2024segmorph}. Biomechanical and deformation based models further rely on detailed anatomical annotations to derive physiological parameters, resulting in substantial labeling burden \cite{sundar2009biomechanically,qin2020biomechanics, gomez2025cascade,yang2024echocardiography}. In contrast, we model cardiac motion using an extremely sparse point tracking strategy that requires only a single annotated template frame for the entire dataset. We transfer this template to initialize landmarks in each video and propagate them over time using a tracking model \cite{karaev2025cotracker3}, producing dense trajectories with minimal supervision.

Foundation models (FMs) have recently become a dominant paradigm in medical image analysis, driven by large scale pretraining and contrastive learning, leading to growing interest in MI detection from Echo \cite{christensen2024vision,kim2025echofm,vukadinovic2025comprehensive}. Recent work has also explored anatomical significance and multi-task learning for explainable MI prediction \cite{peng2025anatomical}. EchoPrime \cite{vukadinovic2025comprehensive}, trained on over 12 million image-text pairs and evaluated across 23 cardiac benchmarks, provides a high capacity visual backbone for routine tasks. However, despite reliable global predictions, our experiments indicate that EchoPrime remains limited for segment level MI localization, particularly in structurally ambiguous regions in the A4C view. This motivates utilizing cardiac motion as a complementary cue to guide visual attention toward pathology relevant regions. 

Motivated by the complementary strengths and inherent limitations of motion and vision based approaches, we propose \textbf{MCF-Net}, a Motion Conditioned Fusion Network for segment level MI localization. MCF-Net adopts a pretrained foundation model as the primary vision encoder and integrates motion guidance through two components: (i) motion-guided soft refinement, which constructs a trajectory-derived Gaussian mask along the tracked myocardial boundary to lead the encoder towards relevant regions, and (ii) motion conditioned multi-view fusion, which uses motion embeddings to gate visual features and applies cross-view attention to aggregate complementary evidence from both views. MCF-Net consistently outperforms all motion only methods (trajectory/motion features only), vision only methods (visual features only), and previous fusion baselines, including strong FMs, achieving up to +8.1\% F1 and +4.8\% accuracy gains for segment MI localization. Our main contributions are: 

\begin{itemize}
    \item We introduce a motion conditioned multi-view localization framework that combines a pretrained echocardiography foundation model with sparse motion priors to enable segment level MI localization across apical views.
    \item We design a motion guided refinement module (soft mask) and a motion conditioned fusion strategy that selectively pushes visual representations toward pathology relevant regions, improving hard myocardial segments (such as basal regions) while preserving strong global appearance cues.
    \item We propose a sparse motion modeling strategy using a single annotated frame with template transfer and point tracking, enabling robust and scalable cardiac motion extraction across views without the need for dense labels.
\end{itemize}

\section{Method}
\begin{figure}[htbp]
    \centering
    \includegraphics[width=\textwidth]{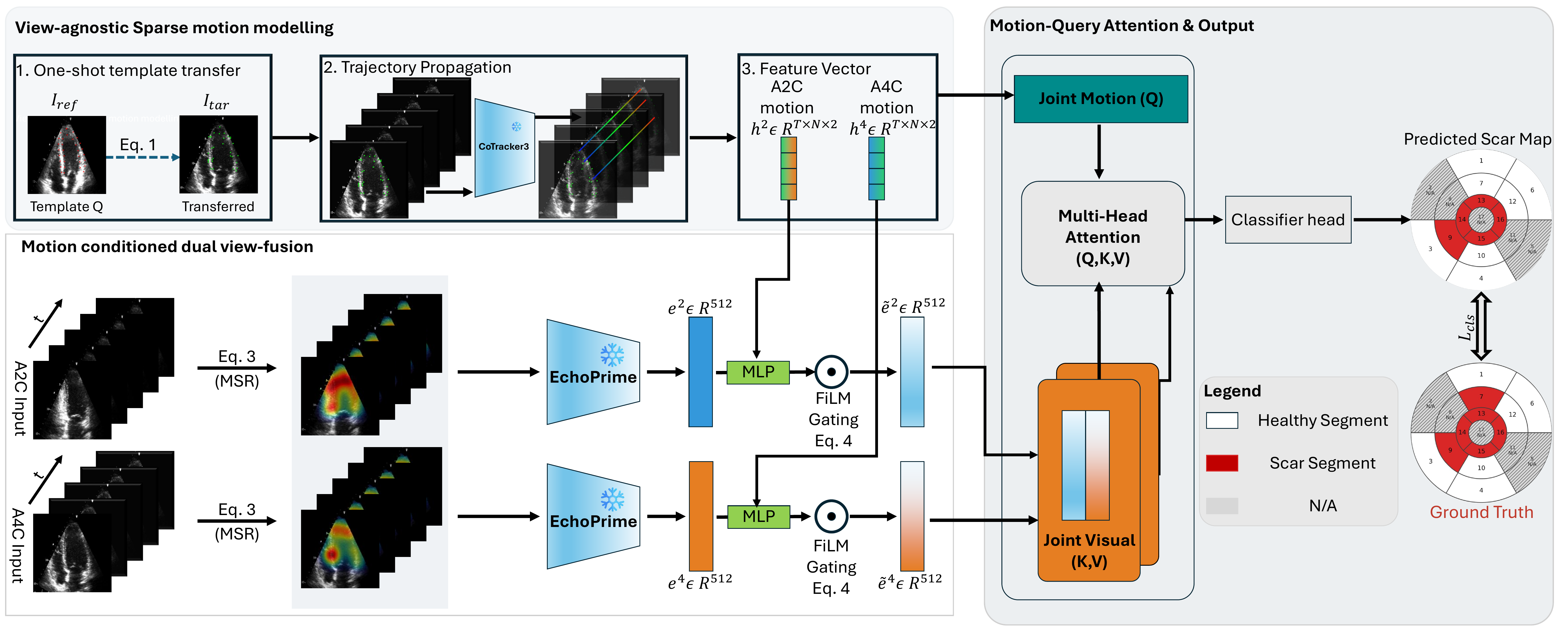}
    \caption{\textbf{Overview of MCF-Net.} (1) \textbf{Sparse Motion Modeling} extracts motion embeddings ($h^2, h^4$) via CoTracker3. (2) \textbf{Dual View Fusion} modulates frozen EchoPrime visual features with motion via FiLM gating (Eq. 4). (3) \textbf{Motion Query Attention} fuses representations using Joint Motion as $Q$ and Visual as $K,V$ to predict 12 segment MI distribution, optimized via $\mathcal{L}_{cls}$.}
    \label{fig:method}
\end{figure}

\noindent Given a paired dual view Echo 
$\mathcal{X}=\big(\mathcal{V}^{(\mathrm{A2C})}, \mathcal{V}^{(\mathrm{A4C})}\big)$,
where each view is represented as a temporal sequence 
$\mathcal{V}^{(v)}=\{ I_t^{(v)} \in \mathbb{R}^{H \times W} \}_{t=1}^{T}$ 
with $v \in \{\mathrm{A2C}, \mathrm{A4C}\}$,
our goal is segment-level MI localization, learning $f_{\theta}:\mathcal{X}\rightarrow \mathbf{z}\in\mathbb{R}^{12}$ with $\mathbf{z}$ the segment-wise logits. As shown in Fig.~\ref{fig:method}, \textbf{MCF-Net} integrates sparse motion priors with foundation appearance features via: (i) sparse motion modeling, (ii) motion-guided refinement, and (iii) motion-conditioned cross-view fusion.

\subsection{Cardiac Motion Modeling from Extremely Sparse Annotation}
To avoid the extensive cost of dense frame by frame human annotation, we extract cardiac motion under extremely sparse supervision. Specifically, our approach requires only one manually annotated frame for the entire dataset: we annotate the canonical landmarks on the first frame of a single reference A4C video and initialize landmarks for all videos via localized template matching followed by point tracking. When switching to A2C view, we directly transfer the A4C template across views using the same matching and tracking pipeline. Importantly, A2C and A4C views are initialized and tracked independently. No explicit point-to-point anatomical correspondence between views is assumed. Formally, we define a canonical landmark set $\mathcal{Q}=\{q_i\}_{i=1}^{N}$ with $N=32$ on the first frame of a reference A4C video $\mathcal{I}_{\mathrm{ref}}$, where the points are sampled along the myocardial wall with approximately uniform allocation across the six view-specific segments. For each landmark $q_i$, we extract a feature patch $\mathcal{P}_i \in \mathbb{R}^{11 \times 11}$ centered at $q_i$. To transfer these landmarks to the first frame of a target video $\mathcal{I}_{\mathrm{tar}}$, we perform localized patch matching within a search window $\Omega \in \mathbb{R}^{41 \times 41}$ centered at the same spatial coordinate $q_i$. Correspondence between anatomical features is established by maximizing the Normalized Cross Correlation (NCC):

\begin{equation}
\mathrm{NCC}(\mathbf{u}) =
\frac{\sum_{\mathbf{x} \in \mathcal{P}_i} 
(\mathcal{I}_{\mathrm{ref}}(\mathbf{x}) - \bar{\mathcal{I}}_{\mathrm{ref}})
(\mathcal{I}_{\mathrm{tar}}(\mathbf{x}+\mathbf{u}) - \bar{\mathcal{I}}_{\mathrm{tar}})}
{\sqrt{
\sum_{\mathbf{x}} (\mathcal{I}_{\mathrm{ref}}(\mathbf{x}) - \bar{\mathcal{I}}_{\mathrm{ref}})^2
\sum_{\mathbf{x}} (\mathcal{I}_{\mathrm{tar}}(\mathbf{x}+\mathbf{u}) - \bar{\mathcal{I}}_{\mathrm{tar}})^2}},
\quad \mathbf{u}=(u_x,u_y)\in\Omega .
\end{equation}
To ensure our method can robustly locate $P=\{p_i(1)\}_{i=1}^{N}$ against acoustic artifacts and structural variations, we employ a high confidence fallback mechanism. The initialized point $p_i$ in $\mathcal{I}_{\mathrm{tar}}$ is defined as follows: 
\begin{equation}
p_i = q_i + \arg\max_{\mathbf{u}\in\Omega}\mathrm{NCC}(\mathbf{u}).
\end{equation}
If $\max_{\mathbf{u}\in\Omega} \mathrm{NCC}(\mathbf{u}) < \tau$, we set $p_i=q_i$, where $\tau=0.5$. This fallback ensures that if no reliable structural match is found (e.g., due to poor echo quality), the coordinate defaults to the atlas prior $q_i$, maintaining a structurally plausible heart shape. The initialized landmarks $\{p_i(1)\}_{i=1}^{N}$ are propagated across $T$ frames using CoTracker3 \cite{karaev2025cotracker3}, producing dense  $\mathbf{P}=\{p_i(t)\}_{t=1,i=1}^{T,N}\in\mathbb{R}^{T\times N\times2}$.

\subsection{Foundation Encoding with Motion Guided Soft Refinement}
While FMs such as EchoPrime provide strong spatial representations, they often lack temporal specificity for infarction detection, where subtle wall thinning and hypokinesia along the myocardial boundary are key biomarkers. To bridge this gap, we introduce \textbf{Motion Guided Soft Refinement (MSR)}, which injects tracked trajectories into the appearance stream. Since $\mathcal{Q}$ is sampled along the myocardial wall with roughly equal points per segment, the resulting trajectories $\{p_i(t)\}_{i=1}^{N}$ induce a segment-aware spatial prior. Therefore we construct a spatial-temporal soft attention mask $M^{(v)}(t,x,y)$ using a Gaussian mixture centered at the landmark locations: 
$M^{(v)}(t,x,y) = \sum_{i=1} ^{N} \exp\!\left(-\frac{\|(x,y)-p_i(t)\|_2^2}{2\sigma^2}\right)$. This mask highlights myocardial regions and is used to softly modulate the input video $\mathcal{V}^{(v)}$ as follows: 
\begin{equation}
 \tilde{\mathcal{V}}^{(v)} = \mathcal{V}^{(v)} \odot \big(1 + \lambda M^{(v)}\big),
 \label{msr}
\end{equation} where $\lambda$ controls attention strength and $\odot$ denotes element wise multiplication. This operation enhances intensity and edge responses around the myocardial wall while largely preserving the original appearance distribution and surrounding context (e.g., the left ventricular cavity and surroundings). The refined video is then fed into the frozen foundation encoder $f_{\phi}$ to extract view level embeddings $e^{(v)}=f_{\phi}\!\left(\tilde{\mathcal{V}}^{(v)}\right)\in\mathbb{R}^{512}$. Overall, this refinement retains the pretrained spatial priors of $f_{\phi}$ while injecting localized motion guidance, improving infarction sensitive feature extraction without altering the encoder parameters.

\subsection{Motion Conditioned Cross View Fusion}
MI localization benefits from complementary evidence across both views, since different AHA segments are best visualized under different orientations. To exploit this complementarity, we propose a motion conditioned fusion module that explicitly couples visual representations with cardiac motion. For each view $v$, the frozen foundation encoder outputs a visual embedding $e^{(v)}\in\mathbb{R}^{512}$, while the tracked motion trajectory $\mathbf{P}^{(v)}\in\mathbb{R}^{T\times N\times 2}$ is summarized by an MLP into a compact motion descriptor $h^{(v)}\in\mathbb{R}^{128}$. We then modulate appearance features via FiLM style conditioning \cite{perez2018film}:
\begin{equation}
\tilde e^{(v)}=\gamma\!\left(h^{(v)}\right)\odot e^{(v)}+\beta\!\left(h^{(v)}\right)\in\mathbb{R}^{512},
\end{equation}
where $\gamma(\cdot)$ and $\beta(\cdot)$ are learned linear mappings. This conditioning enables motion abnormalities (e.g., reduced contraction patterns) to re-weight the visual embedding in a view-specific manner, promoting infarction-relevant cues while suppressing non informative variations. To fuse across two views, we concatenate the modulated embeddings $V=[\tilde e^{(\mathrm{A2C})},\,\tilde e^{(\mathrm{A4C})}]$ and form a joint motion query $q=[\,h^{(\mathrm{A2C})};\,h^{(\mathrm{A4C})}\,]$. Cross view fusion is performed using multi-head attention, $ z=\mathrm{MHA}(q, V, V)\in\mathbb{R}^{512},$ so that pooling across views is explicitly guided by patient-specific motion. Finally, the fused representation is mapped to segment-wise infarction probabilities $\hat y=\sigma(W_o z)\in(0,1)^{12}$. The training objective is optimized by segment-wise classification loss $L_{\mathrm{cls}} = L_{\mathrm{BCE}}(y,\hat{y}).$ Overall, the proposed fusion couples motion-informed inter-view modulation with attention-based cross-view aggregation, enabling view-complementary and pathologically-informed representations that improve segment level MI localization.

\section{Experiments}

\subsection{Materials}

\subsubsection{Dataset}
This study uses the HMC-QU dataset \cite{degerli2024early}, the only publicly available Echo dataset for MI detection to date. HMC-QU comprises 162 A4C and 160 A2C Echo videos, with paired A2C and A4C views available for 160 patients. Following the standardized AHA 17-segment model \cite{cerqueira2002standardized}, MI presence is annotated for the six segments visible in each view. We randomly partitioned the data into 112/16/32 patients for training, validation, and testing. All experiments were repeated three times, with performance reported as segment-level averages.

\subsubsection{Implementation} 
All experiments were conducted on a workstation equipped with an Intel i7-13700 CPU and an NVIDIA RTX A4000 GPU. Echo videos were preprocessed by resizing frames to 224 × 224 pixels and uniformly downsampled to 16 frames to meet the input requirements of EchoPrime. The proposed framework was implemented in PyTorch. Model training employed the AdamW optimizer with a learning rate of $1\times10^{-3}$ and a weight decay of $1\times10^{-4}$.
 All models were trained for 50 epochs with a batch size of 16.

\subsection{Results}
We evaluate segment-level MI localization using threshold-free (AUROC, PR-AUC) and threshold-dependent (F1, ACC) metrics, averaged over three runs. MCF-Net is compared against seven baselines spanning motion-only \cite{kiranyaz2020left,degerli2024early}, vision-only \cite{szegedy2016rethinking,tran2018closer,vukadinovic2025comprehensive}, and fusion-based methods, including naïve EchoPrime fusion and EchoAna \cite{yang2024echocardiography}. All models operate on full echo videos, enabling systematic comparison of motion priors, foundation representations, and fusion strategies.
\begin{table}[ht]
\centering
\caption{Quantitative comparison of segment level MI localization. Experiments are run three times with different seeds. Results are reported as mean $\pm$ standard deviation. 
Best results are \textbf{bold}. 
PR-AUC denotes the area under the precision--recall curve.}
\label{tab:sota}
\fontsize{8}{9}\selectfont
\setlength{\tabcolsep}{3.5pt}
\begin{tabular}{lccccc c}
\toprule
\textbf{Methods} & \textbf{Motion} & \textbf{Vision} 
& \textbf{AUROC} & \textbf{PR-AUC} & \textbf{F1-Score} & \textbf{Accuracy} \\
\midrule

\rowcolor{motion}
APs \cite{kiranyaz2020left}
& \cmark & \xmark 
& -- & -- 
& 57.7 $\pm$ 5.8 & 79.0 $\pm$ 2.6 \\

\rowcolor{motion}
APsML \cite{degerli2024early}
& \cmark & \xmark 
& 83.0 $\pm$ 2.1 & 70.4 $\pm$ 7.2 
& 59.9 $\pm$ 6.0 & 79.8 $\pm$ 1.7 \\

\midrule

\rowcolor{vision}
InceptionV3 \cite{szegedy2016rethinking}
& \xmark & \cmark 
& 61.1 $\pm$ 8.0 & 47.2 $\pm$ 12.0 
& 27.0 $\pm$ 2.8 & 61.5 $\pm$ 1.1 \\

\rowcolor{vision}
R(2+1)D \cite{tran2018closer}
& \xmark & \cmark 
& 77.3 $\pm$ 1.7 & 58.2 $\pm$ 3.9 
& 45.6 $\pm$ 9.8 & 73.3 $\pm$ 5.4 \\

\rowcolor{vision}
EchoPrime \cite{vukadinovic2025comprehensive}
& \xmark & \cmark 
& {84.9 $\pm$ 2.6} & {72.7 $\pm$ 6.2} 
& {64.3 $\pm$ 3.8} & 80.1 $\pm$ 2.1 \\

\midrule

\rowcolor{fusion}
EchoPrimeNF \cite{vukadinovic2025comprehensive}
& \cmark & \cmark 
& 84.4 $\pm$ 1.8 & 70.5 $\pm$ 6.2 
& 59.3 $\pm$ 5.6 & 78.8 $\pm$ 1.7 \\

\rowcolor{fusion}
EchoAna \cite{yang2024echocardiography}
& \cmark & \cmark 
& 84.6 $\pm$ 1.9 & 72.1 $\pm$ 6.5 
& 61.7 $\pm$ 4.6 & {80.9 $\pm$ 2.8} \\

\rowcolor{fusion}
\textbf{Ours} 
& \cmark & \cmark 
& \textbf{87.6 $\pm$ 2.4} & \textbf{74.7 $\pm$ 5.6} 
& \textbf{72.4 $\pm$ 3.8} & \textbf{84.9 $\pm$ 2.3} \\

\bottomrule
\end{tabular}
\end{table}


\subsubsection{Comparison Studies}
\begin{figure}[ht]
    \centering
    \includegraphics[width=\textwidth]{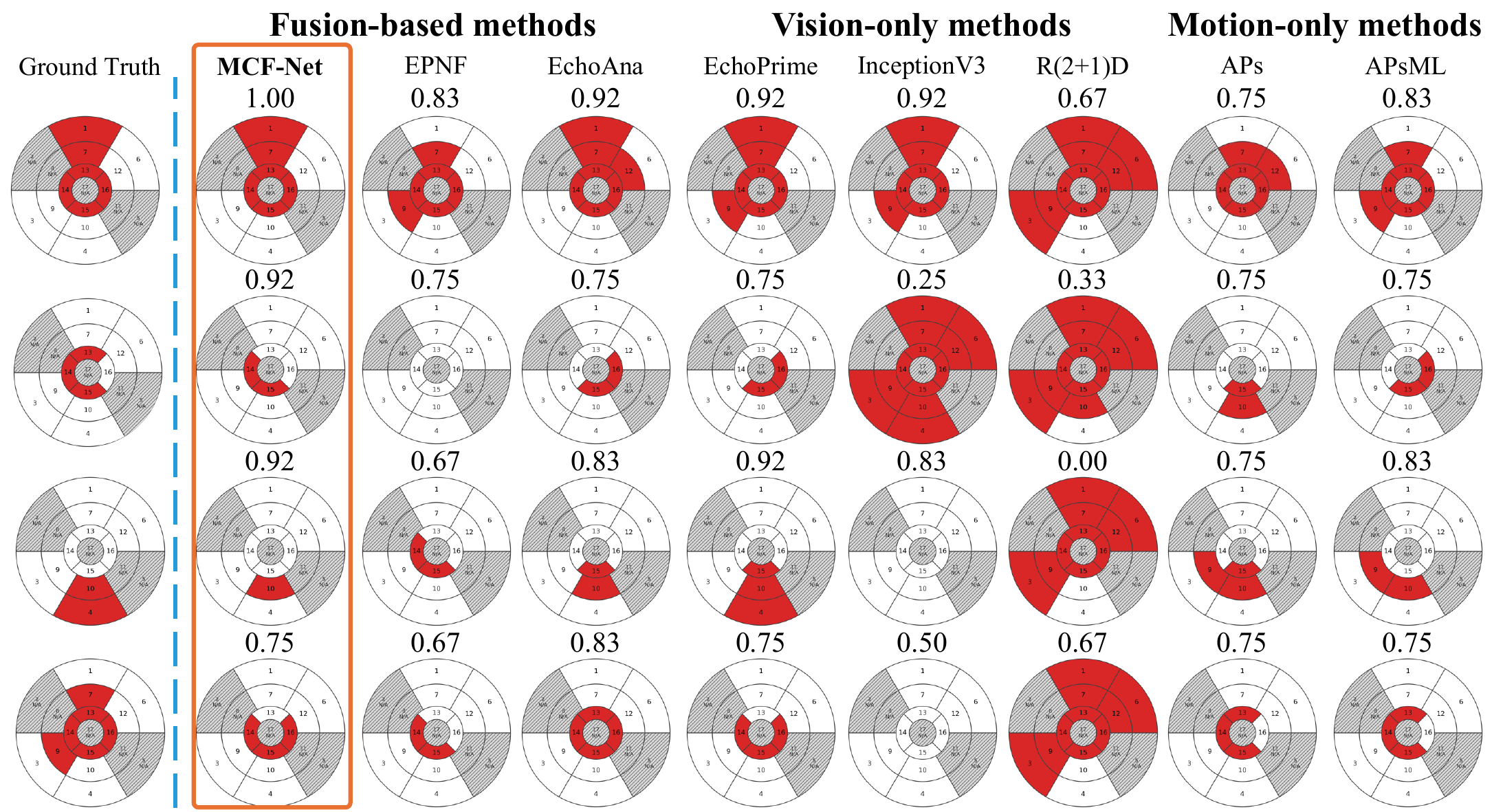}
    \caption{
    \textbf{Qualitative comparison on AHA bullseye plots.} Each ``segment'' denotes a standardized \emph{anatomical region} of the left ventricle (apical/mid/basal; anterior/inferior, etc.), i.e., region-level labeling rather than pixel-wise segmentation. \textbf{Red, white, gray} denote infarct, healthy, and N/A regions, respectively. Rows show representative test cases selected by accuracy quartiles (0.75: lower, 0.92: median, 1.00: upper), comparing Ground Truth (GT) with \textbf{MCF-Net (Ours)} and baselines grouped by modality (Fusion, Vision, Motion). Numbers indicate accuracy.
    }
    \label{fig:results}
\end{figure}
Table~\ref{tab:sota} shows that MCF-Net achieves the best performance across all metrics for segment level MI localization (AUROC $87.6$, PR-AUC $74.7$, F1 $72.4$, ACC $84.9$). Improvements are consistent in both threshold-free (AUROC/PR-AUC) and threshold-dependent (F1/ACC) metrics, indicating gains in discrimination and region level decision quality. \textbf{Motion only methods} (APs, APsML) achieve reasonable overall accuracy ($\sim80$) but lack spatial precision, resulting in limited segment level localization (APsML: PR-AUC $70.4\pm7.2$, F1 $59.9\pm6.0$). \textbf{Vision only models} show that standard backbones struggle (InceptionV3 F1 $27.0\pm2.8$, R(2+1)D F1 $45.6\pm9.8$), while the pretrained foundation model EchoPrime substantially improves performance (AUROC $84.9\pm2.6$, F1 $64.3\pm3.8$), highlighting the benefit of large scale pretraining. \textbf{Fusion methods prove critical.} Naïve fusion (EchoPrimeNF) degrades performance relative to EchoPrime (F1 $59.3\pm5.6$ vs.\ $64.3\pm3.8$), suggesting misaligned motion-vision interaction. In contrast, our motion-conditioned fusion consistently improves all metrics, demonstrating the benefit of conditioning visual representations before cross view aggregation. In addition, paired patient-level bootstrap analysis confirms the F1 improvement of MCF-Net over EchoPrime (Delta=0.08, 95\% CI=[0.01,0.15], p<0.05), indicating that the observed performance gain is statistically significant. Qualitative results in Fig.~\ref{fig:results} further support these findings. Across representative cases spanning lower to upper performance quartiles, MCF-Net produces compact and anatomically coherent infarct regions. Vision only methods often overestimate lesions (diffuse false positives), while motion only approaches generate fragmented patterns. Alternative fusion strategies may shift activation to adjacent segments. In contrast, MCF-Net more accurately recovers infarct extent and topology, consistent with its quantitative gains.

\subsubsection{Ablation Study}
\begin{table}[ht]
\centering
\caption{Ablation study of Motion Conditioned Fusion (MCF) and Motion Guided Soft Refinement (MSR). Results are reported as mean $\pm$ standard deviation.}
\label{tab:ablation}
\fontsize{8}{9}\selectfont
\setlength{\tabcolsep}{3.5pt}
\begin{tabular}{lccccc c}
\toprule
\textbf{Variant} & \textbf{MCF} & \textbf{MSR} 
& \textbf{AUROC} & \textbf{PR-AUC} & \textbf{F1-Score} & \textbf{Accuracy} \\
\midrule
\rowcolor{motion}
Baseline           
& \xmark & \xmark 
& 84.9 $\pm$ 2.6 & 72.7 $\pm$ 6.2 & 64.3 $\pm$ 3.8 & 80.1 $\pm$ 2.1 \\
\midrule

\rowcolor{vision}
Baseline + MSR     
& \xmark & \cmark 
& 86.9 $\pm$ 2.8 & 71.9 $\pm$ 6.7 & 66.8 $\pm$ 3.6 & 82.4 $\pm$ 3.6 \\

\rowcolor{vision}
Baseline + MCF     
& \cmark & \xmark 
& 87.2 $\pm$ 2.5 & \textbf{74.7 $\pm$ 5.8} & 70.3 $\pm$ 4.8 & 83.8 $\pm$ 3.3 \\
\midrule
\rowcolor{fusion}
\textbf{Full Model} 
& \cmark & \cmark 
& \textbf{87.6 $\pm$ 2.4} & 74.6 $\pm$ 5.6 
& \textbf{72.4 $\pm$ 3.8} & \textbf{84.9 $\pm$ 2.3} \\
\bottomrule
\end{tabular}
\end{table}
As shown in Table~\ref{tab:ablation}, the baseline model (EchoPrime encoder + linear head) achieves strong overall performance (AUROC $84.9\pm2.6$, F1 $64.3\pm3.8$), reflecting the benefit of large scale pretraining for global representations, yet its gains are less pronounced when segment level MI localization is required. Adding MSR yields a modest but consistent improvement (AUROC $86.9\pm2.8$, F1 $66.8\pm3.6$), suggesting that the soft mask mainly sharpens the decision boundary by steering features towards the myocardial region. Incorporating MCF provides a larger boost (F1 $70.3\pm4.8$, ACC $83.8\pm3.3$, PR-AUC $74.7\pm5.8$), indicating increased model capacity to capture motion appearance interactions that are critical for segment level reasoning. Combining both modules achieves the best results (AUROC $87.6\pm2.4$, F1 $72.4\pm3.8$, ACC $84.9\pm2.3$), confirming their complementarity: MCF strengthens cross domain fusion, while MSR refines spatial emphasis to improve separability and localization robustness.

\subsubsection{The Effect of MSR Strength $\lambda$}
\begin{figure}[ht]
  \centering
  \includegraphics[width=\linewidth]{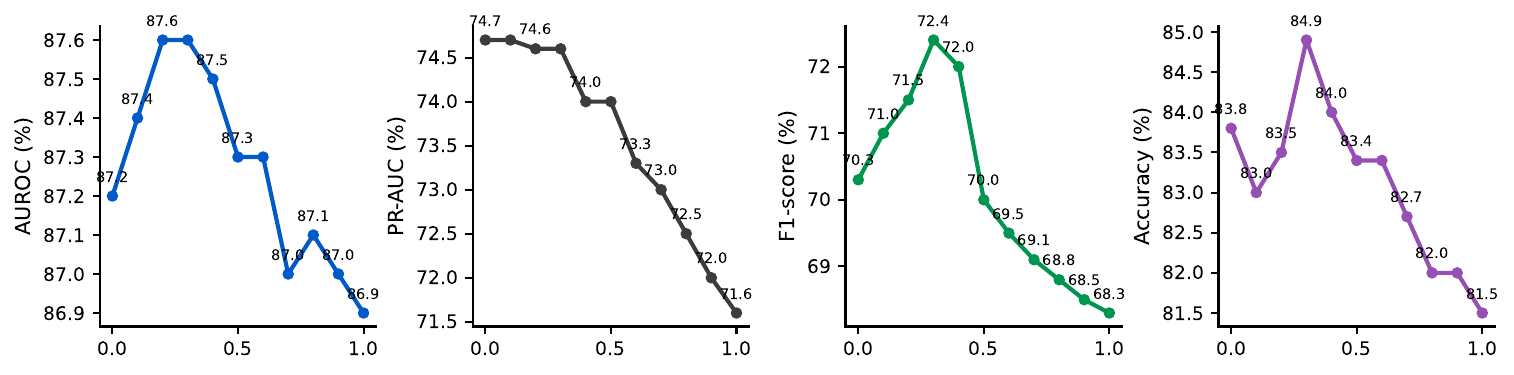}
  \caption{Sensitivity analysis of the refinement strength $\lambda$ on the test set (higher $\lambda$ strengthens masking). From left to right: AUROC, PR-AUC, F1-score, and accuracy.}
  \label{fig:lambda_sensitivity}
\end{figure}

We analyze the impact of $\lambda$ in Equation \ref{msr}, which scales the motion derived soft mask. As shown in Fig. \ref{fig:lambda_sensitivity}, performance follows a clear trend, with a moderate refinement strength ($\lambda = 0.3$) yielding the best overall performance (AUROC 87.6, F1 72.4, Acc 84.9). A stable high performance region is observed for $\lambda \in [0.2,0.4]$, indicating robustness to mild hyperparameter variation. However, performance consistently degrades when $\lambda \ge 0.5$, with F1 decreasing to $68.3$ at $\lambda=1.0$, falling below the baseline. This suggests that excessive masking can distort the original Echo, lead to artefactual enhancements and ultimately degrade localization reliability. Notably, while Accuracy and F1 improve significantly at moderate $\lambda$, PR-AUC remains relatively stable, implying that soft refinement primarily improves decision boundaries rather than ranking quality. These findings support $\lambda=0.3$ as an optimal trade-off between motion enhancement and visual fidelity.

\section{Conclusion}
In this work, we present MCF-Net, a Motion Conditioned Fusion Network for segment level myocardial infarction localization from dual view echocardiography. Our framework utilizes a pretrained foundation model to extract rich visual representations and introduces Motion Guided Soft Refinement to focus feature extraction on infarction-relevant regions. Motion Conditioned Fusion modulates appearance features using view-specific motion and enables cross-view interaction between motion dynamics and visual semantics, improving fine grained localization. Experimental results validate that sparse physiological motion priors can further enhance strong foundation models such as EchoPrime, providing a practical approach for MI localization from routine echocardiography. Future work will investigate larger multi-center validation and systematic failure-case analysis.

\begin{credits}

\subsubsection{\ackname}
The authors acknowledge the use of computing cluster resources provided by the Institute of Biomedical Engineering, Department of Engineering Science, University of Oxford.

\subsubsection{\discintname}

The authors declare no competing interests.

\end{credits}

\bibliographystyle{splncs04}
\bibliography{paper-3435}

@incollection{zeidan2024myocardial,
  title={Myocardial Infarctions in Developing Countries},
  author={Zeidan, Rouba Karen and Farah, Rita},
  booktitle={Handbook of Medical and Health Sciences in Developing Countries: Education, Practice, and Research},
  pages={1--30},
  year={2024},
  publisher={Springer}
}

@article{porter2018clinical,
  title={Clinical applications of ultrasonic enhancing agents in echocardiography: 2018 American Society of Echocardiography guidelines update},
  author={Porter, Thomas R and Mulvagh, Sharon L and Abdelmoneim, Sahar S and Becher, Harald and Belcik, J Todd and Bierig, Michelle and Choy, Jonathan and Gaibazzi, Nicola and Gillam, Linda D and Janardhanan, Rajesh and others},
  journal={Journal of the American Society of Echocardiography},
  volume={31},
  number={3},
  pages={241--274},
  year={2018},
  publisher={Elsevier}
}

@article{suhling2005myocardial,
  title={Myocardial motion analysis from B-mode echocardiograms},
  author={Suhling, Michael and Arigovindan, Muthuvel and Jansen, Christian and Hunziker, Patrick and Unser, Michael},
  journal={IEEE Transactions on image processing},
  volume={14},
  number={4},
  pages={525--536},
  year={2005},
  publisher={IEEE}
}

@article{ortiz2023optical,
  title={Optical Flow-Guided Cine MRI Segmentation With Learned Corrections},
  author={Ortiz-Gonzalez, Antonio and Kobler, Erich and Simon, Stefan and Bischoff, Leon and Nowak, Sebastian and Isaak, Alexander and Block, Wolfgang and Sprinkart, Alois M and Attenberger, Ulrike and Luetkens, Julian A and others},
  journal={IEEE Transactions on Medical Imaging},
  volume={43},
  number={3},
  pages={940--953},
  year={2023},
  publisher={IEEE}
}

@article{balakrishnan2019voxelmorph,
  title={Voxelmorph: a learning framework for deformable medical image registration},
  author={Balakrishnan, Guha and Zhao, Amy and Sabuncu, Mert R and Guttag, John and Dalca, Adrian V},
  journal={IEEE transactions on medical imaging},
  volume={38},
  number={8},
  pages={1788--1800},
  year={2019},
  publisher={IEEE}
}

@inproceedings{yang2025contrast,
  title={Contrast-Free Myocardial Scar Segmentation in Cine MRI using Motion and Texture Fusion},
  author={Yang, Guang and Chen, Jingkun and Sheng, Xicheng and Yang, Shan and Zhuang, Xiahai and Raman, Betty and Li, Lei and Grau, Vicente},
  booktitle={2025 IEEE 22nd International Symposium on Biomedical Imaging (ISBI)},
  pages={1--5},
  year={2025},
  organization={IEEE}
}

@article{bi2024segmorph,
  title={Segmorph: concurrent motion estimation and segmentation for cardiac mri sequences},
  author={Bi, Ning and Zakeri, Arezoo and Xia, Yan and Cheng, Nina and Taylor, Zeike A and Frangi, Alejandro F and Gooya, Ali},
  journal={IEEE transactions on medical imaging},
  year={2024},
  publisher={IEEE}
}

@inproceedings{sundar2009biomechanically,
  title={Biomechanically-constrained 4D estimation of myocardial motion},
  author={Sundar, Hari and Davatzikos, Christos and Biros, George},
  booktitle={International Conference on Medical Image Computing and Computer-Assisted Intervention},
  pages={257--265},
  year={2009},
  organization={Springer}
}

@inproceedings{qin2020biomechanics,
  title={Biomechanics-informed neural networks for myocardial motion tracking in MRI},
  author={Qin, Chen and Wang, Shuo and Chen, Chen and Qiu, Huaqi and Bai, Wenjia and Rueckert, Daniel},
  booktitle={International conference on medical image computing and computer-assisted intervention},
  pages={296--306},
  year={2020},
  organization={Springer}
}

@article{yang2024echocardiography,
  title={Echocardiography Analysis with Deep Learning using Priors: Multi-centric Evaluation of Generalisation},
  author={Yang, Yingyu and Rocher, Marie and Moceri, Pamela and Sermesant, Maxime and others},
  journal={Machine Learning for Biomedical Imaging},
  volume={2},
  number={November 2024 issue},
  pages={2293--2325},
  year={2024}
}

@article{christensen2024vision,
  title={Vision--language foundation model for echocardiogram interpretation},
  author={Christensen, Matthew and Vukadinovic, Milos and Yuan, Neal and Ouyang, David},
  journal={Nature Medicine},
  volume={30},
  number={5},
  pages={1481--1488},
  year={2024},
  publisher={Nature Publishing Group US New York}
}

@article{kim2025echofm,
  title={EchoFM: Foundation model for generalizable echocardiogram analysis},
  author={Kim, Sekeun and Jin, Pengfei and Song, Sifan and Chen, Cheng and Li, Yiwei and Ren, Hui and Li, Xiang and Liu, Tianming and Li, Quanzheng},
  journal={IEEE transactions on medical imaging},
  year={2025},
  publisher={IEEE}
}

@article{vukadinovic2025comprehensive,
  title={Comprehensive echocardiogram evaluation with view primed vision language AI},
  author={Vukadinovic, Milos and Chiu, I-Min and Tang, Xiu and Yuan, Neal and Chen, Tien-Yu and Cheng, Paul and Li, Debiao and Cheng, Susan and He, Bryan and Ouyang, David},
  journal={Nature},
  pages={1--3},
  year={2025},
  publisher={Nature Publishing Group UK London}
}

@inproceedings{karaev2025cotracker3,
  title={Cotracker3: Simpler and better point tracking by pseudo-labelling real videos},
  author={Karaev, Nikita and Makarov, Yuri and Wang, Jianyuan and Neverova, Natalia and Vedaldi, Andrea and Rupprecht, Christian},
  booktitle={Proceedings of the IEEE/CVF International Conference on Computer Vision},
  pages={6013--6022},
  year={2025}
}

@article{degerli2024early,
  title={Early myocardial infarction detection over multi-view echocardiography},
  author={Degerli, Aysen and Kiranyaz, Serkan and Hamid, Tahir and Mazhar, Rashid and Gabbouj, Moncef},
  journal={Biomedical Signal Processing and Control},
  volume={87},
  pages={105448},
  year={2024},
  publisher={Elsevier}
}

@article{kiranyaz2020left,
  title={Left ventricular wall motion estimation by active polynomials for acute myocardial infarction detection},
  author={Kiranyaz, Serkan and Degerli, Aysen and Hamid, Tahir and Mazhar, Rashid and Ahmed, Rayyan El Fadil and Abouhasera, Rayaan and Zabihi, Morteza and Malik, Junaid and Hamila, Ridha and Gabbouj, Moncef},
  journal={IEEE Access},
  volume={8},
  pages={210301--210317},
  year={2020},
  publisher={IEEE}
}

@article{cerqueira2002standardized,
  title={Standardized myocardial segmentation and nomenclature for tomographic imaging of the heart: a statement for healthcare professionals from the Cardiac Imaging Committee of the Council on Clinical Cardiology of the American Heart Association},
  author={Cerqueira, MD and Weissman, NJ and Dilsizian, V and Jacobs, AK},
  journal={Circulation},
  volume={105},
  number={4},
  pages={539--542},
  year={2002}
}

@inproceedings{tran2018closer,
  title={A closer look at spatiotemporal convolutions for action recognition},
  author={Tran, Du and Wang, Heng and Torresani, Lorenzo and Ray, Jamie and LeCun, Yann and Paluri, Manohar},
  booktitle={Proceedings of the IEEE conference on Computer Vision and Pattern Recognition},
  pages={6450--6459},
  year={2018}
}

@inproceedings{perez2018film,
  title={Film: Visual reasoning with a general conditioning layer},
  author={Perez, Ethan and Strub, Florian and De Vries, Harm and Dumoulin, Vincent and Courville, Aaron},
  booktitle={Proceedings of the AAAI conference on artificial intelligence},
  volume={32},
  year={2018}
}

@inproceedings{peng2025anatomical,
  title={An Anatomical Significance-Aware Architecture for Explainable Myocardial Infarction Prediction via Multi-task Learning},
  author={Peng, Jiachuan and Beetz, Marcel and Banerjee, Abhirup and Chen, Min and Grau, Vicente},
  booktitle={International Conference on Medical Image Computing and Computer-Assisted Intervention},
  pages={13--23},
  year={2025},
  organization={Springer}
}

@inproceedings{lyu2025personalized,
  title={Personalized 3D Myocardial Infarct Geometry Reconstruction from Cine MRI with Explicit Cardiac Motion Modeling},
  author={Lyu, Yilin and Yang, Fan and Liu, Xiaoyue and Jiang, Zichen and Dillon, Joshua and Zhao, Debbie and Nash, Martyn and Mauger, Charlene and Young, Alistair and Sia, Ching-Hui and others},
  booktitle={International Workshop on Digital Twin for Healthcare},
  pages={1--11},
  year={2025},
  organization={Springer}
}

@article{gomez2025cascade,
  title={A cascade approach for the early detection and localization of myocardial infarction in 2D-echocardiography},
  author={Gomez, Carolina and Letizia, Annalisa and Tufano, Vincenza and Molinari, Filippo and Salvi, Massimo},
  journal={Medical Engineering \& Physics},
  volume={143},
  number={1},
  pages={104400},
  year={2025},
  publisher={IOP Publishing}
}

@inproceedings{szegedy2016rethinking,
  title={Rethinking the inception architecture for computer vision},
  author={Szegedy, Christian and Vanhoucke, Vincent and Ioffe, Sergey and Shlens, Jon and Wojna, Zbigniew},
  booktitle={Proceedings of the IEEE conference on computer vision and pattern recognition},
  pages={2818--2826},
  year={2016}
}

\end{document}